\documentclass[10pt,twocolumn,letterpaper]{article}

\usepackage[pagenumbers]{cvpr} 

\usepackage[utf8]{inputenc} 
\usepackage[T1]{fontenc} 
\usepackage{times}
\usepackage{floatrow}
\usepackage{comment}
\usepackage{bbding}
\usepackage{placeins}
\usepackage[table]{xcolor}
\usepackage{colortbl}
\usepackage{tabularx}
\usepackage{adjustbox}
\usepackage{xspace}
\usepackage[acronym]{glossaries}
\usepackage{wrapfig}
\usepackage{algpseudocode}
\usepackage{mwe}
\usepackage{subcaption}
\usepackage{soul} 
\usepackage{bm}
\usepackage{amssymb} 
\usepackage{microtype}
\usepackage{textcomp}
\usepackage{algorithm}
\usepackage{amsmath}
\usepackage{float}
\floatstyle{plaintop}
\restylefloat{table}

\usepackage{graphicx}
\usepackage{multirow}
\usepackage{booktabs}
\usepackage{array}
\usepackage{gensymb}
\usepackage{physics}
\usepackage{makecell}
\usepackage{pifont}
\usepackage{enumitem}

\definecolor{mygray}{gray}{0.94}
\definecolor{myred}{RGB}{196,15,15}
\definecolor{myblue}{RGB}{33,95,154}
\definecolor{mygreen}{RGB}{57,158,163}
\definecolor{codegreen}{rgb}{0,0.6,0}
\definecolor{codegray}{rgb}{0.5,0.5,0.5}
\definecolor{codepurple}{rgb}{0.58,0,0.82}
\definecolor{backcolour}{rgb}{0.95,0.95,0.92}
\definecolor{mediumtealblue}{rgb}{0.0, 0.33, 0.71}
\definecolor{darkpastelgreen}{rgb}{0.01, 0.75, 0.24}
\definecolor{azure}{rgb}{0, 0.325, 0.420}
\definecolor{Red}{rgb}{1, 0.2, 0.1}
\definecolor{Green}{rgb}{0.1, 1, 0.1}

\usepackage{steinmetz}

\definecolor{cvprblue}{rgb}{0.21,0.49,0.74}
\definecolor{refred}{rgb}{0.8,0.2,0.2}
\definecolor{revblue}{rgb}{0.0, 0.0, 1.0}
\definecolor{revorange}{rgb}{1.0, 0.5, 0.0}
\definecolor{revmagenta}{rgb}{0.8, 0.0, 0.4}

\newcommand{\tref}[1]{\textcolor{black}{Tab.~\ref{#1}}}

\usepackage[pagebackref,breaklinks,colorlinks,allcolors=myblue]{hyperref}

\newcommand{\no}{{\textcolor{myred}{\ding{55}}}}
\newcommand{\ye}{{\textcolor{mygreen}{\ding{51}}}}
\newcommand{\na}{{\textcolor[rgb]{0,0,0.8}{\ding{52}\rotatebox[origin=c]{-9.2}{\kern-0.7em\ding{55}}}}}
\newcommand{\highlightcell}[1]{ \textit{\textbf{\textcolor{azure}{#1}}} }

\usepackage[capitalize]{cleveref}

\newcommand{\name}{\textsc{Diff4Splat}}

\newcommand\nnfootnote[1]{%
\begin{NoHyper}
 \renewcommand\thefootnote{}\footnote{#1}%
\addtocounter{footnote}{-1}%
\end{NoHyper}
}

\def\paperID{37198} 
\def\confName{CVPR}
\def\confYear{2026}

\title{\textsc{Diff4Splat}: Repurposing Video Diffusion Models \\ for Dynamic Scene Generation}

\author{
Panwang Pan$^{*\ddagger}$, Chenguo Lin$^{*}$, Chenxin Li, Jingjing Zhao, Yuchen Lin, \\
Haopeng Li, Yunlong Lin, Kairun Wen, Yixuan Yuan, Yadong MU \\
Peking University, Xiamen University, CUHK, Carnegie Mellon University \\
\nnfootnote{$^*$: Equal contribution; $\dagger$: Project lead; $\ddagger$: Corresponding author. Correspondence: \texttt{paulpanwang@gmail.com}}
\textbf{\url{https://paulpanwang.github.io/Diff4Splat}}
}

\begin{document}
\maketitle

\nnfootnote{$^*$: Equal contribution; $\ddagger$: Corresponding author.}


\begin{abstract}
We introduce \name{}, a feed-forward framework for dynamic scene generation from a single image. Our method synergizes the powerful generative priors of video diffusion models with geometric and motion constraints learned from a large-scale 4D dataset. Given a single image, a camera trajectory, and an optional text prompt, our model directly predicts a dynamic scene represented by a deformable 3D Gaussian field. This approach captures appearance, geometry, and motion in a single pass, eliminating the need for test-time optimization or post-hoc processing. At the core of our framework is a video latent transformer that enhances existing video diffusion models, enabling them to jointly model spatio-temporal dependencies and predict 3D Gaussian Primitives over time. Supervised by objectives targeting appearance fidelity, geometric accuracy, and motion consistency, \name{} generates high-fidelity dynamic scenes within 30 seconds. We demonstrate the effectiveness of \name{} across video generation, novel view synthesis, and geometry extraction, where it matches or surpasses optimization-based methods for dynamic scene synthesis while being significantly more efficient.

\end{abstract}

\section{Introduction}\label{sec:intro}
The generation of dynamic 3D scenes from a single image is a grand challenge in computer vision, with transformative potential for immersive content creation, robotics, and simulation. Current paradigms, however, are caught in a fundamental dilemma. One path involves multi-stage pipelines that first generate a video and then apply 3D reconstruction~\citep{kerbl20233d,yang2024deformable}. These methods are slow, error-prone, and lack end-to-end control. The alternative, direct feed-forward generation, has been largely confined to producing 2D video frames~\citep{ren2025gen3c,bahmani2024ac3d} or static 3D scenes~\citep{liang2024wonderland}, failing to capture the explicit, dynamic 3D geometry essential for true 4D representation. This impasse reveals a critical gap: the absence of a framework that can directly and efficiently synthesize an explicit, controllable scene representation.

\begin{figure*}[!t]
      \begin{center}
 \includegraphics[width=0.85\textwidth]{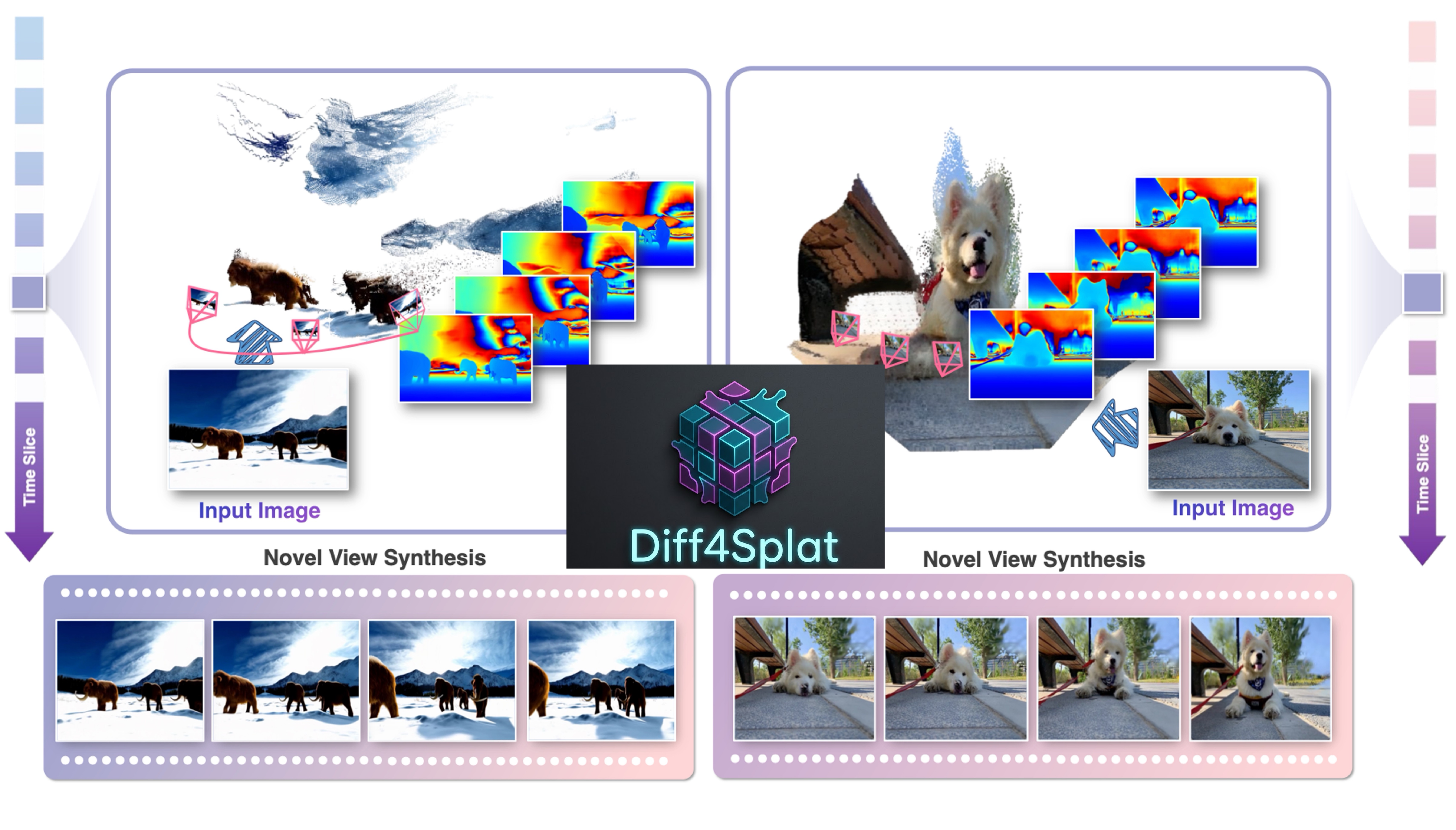}
      \end{center}
      \vspace{-0.3cm}
      \caption{Given a single image, a specified camera trajectory, and an optional text prompt, our diffusion-based framework directly generates a \textbf{deformable 3D Gaussian field without test-time optimization}. The resulting representation supports diverse applications, including video generation, depth rendering, and novel view synthesis, enabling real-time rendering of dynamic scenes and interactive virtual exploration.}
      \label{fig:teaser}
\end{figure*}


To address this, we introduce \name{}, a novel paradigm for dynamic 3D generation that unifies a diffusion backbone with a deformable 3D Gaussian field representation~\cite{deformable3dgaussian} within a single, end-to-end trainable model. As illustrated in Fig.~\ref{fig:teaser}, our framework directly generates a complete 4D representation in a single forward pass, obviating the need for per-scene optimization. The core of our method is a Video Latent Transformer, an architecture engineered to bridge the representational gap between 2D spatio-temporal features and 4D dynamic scenes. This transformer interprets latent features from the diffusion model as a dynamic point cloud, conditioned on camera and temporal embeddings, from which a lightweight head regresses the parameters of the deformable 3D Gaussians. This design compels the diffusion model to learn a powerful prior over explicit 3D geometry and motion, marking a significant departure from prior work~\cite{liang2024wonderland}. To facilitate training, we developed a large-scale data pipeline to annotate real-world videos~\cite{RealEstate10k,jin2024stereo4d} with the metric-scale 4D labels required for supervision. A critical bottleneck in prior work is the exorbitant computational cost of producing explicit 4D representations. Existing methods are hamstrung by a multi-stage process, where a generated video~\citep{zhao2024genxd,ren2025gen3c,bahmani2024ac3d} must undergo a costly, per-scene optimization to be lifted into 3D. The scale of this inefficiency is staggering: methods like DimensionX~\citep{sun2024dimensionx} demand a prohibitive several GPU hours for a single video, while even cutting-edge techniques like Mosca~\citep{lei2024mosca} require half an hour. This renders them impractical for nearly all real-world applications. Our work directly confronts this challenge. By unifying generation and representation into a single feed-forward pass, \name{} collapses this entire pipeline into approximately \textbf{30 seconds}. This leads to a fundamental leap in efficiency, \textbf{a 60-fold acceleration} compared to state-of-the-art optimization, making dynamic 3D scene generation truly practical for the first time.

Our contributions can be summarized as follows:
\begin{itemize}

    \item We introduce a new paradigm for dynamic 3D scene generation, where a diffusion model is trained to directly synthesize a deformable 3D Gaussian field in a single forward pass, resolving the tension between feed-forward efficiency and explicit 3D representation.
    \item We design a novel Video Latent Transformer architecture that bridges the representational gap between 2D latent features and a structured 4D representation, enabling the direct prediction of a deformable 3D Gaussian field.
    \item To enable our method, we construct a large-scale 4D dataset with metric-scale geometry and motion annotations, which will be released to facilitate future research.
    \item Extensive experiments validate that our unified approach produces high-fidelity dynamic 3D scenes from a single image, outperforming complex two-stage pipelines in both quality and efficiency.
\end{itemize}
\vspace{-0.2cm}
\section{Related Work}\label{sec:related}

\paragraph{Video Diffusion Models}
Video diffusion models~\citep{ho2022video} have demonstrated a remarkable capacity for generating high-quality, temporally coherent videos. Fine-grained control is typically achieved by adapting conditional image synthesis strategies~\citep{zhang2023adding,mou2024t2i,li2023gligen} to the video domain, incorporating diverse signals such as RGB images~\citep{blattmann2023svd,xing2023dynamicrafter,xing2024tooncrafter}, depth maps~\citep{xing2024make,esser2023structure}, motion trajectories~\citep{yin2023dragnuwa,niu2024mofa}, and semantic maps~\citep{peruzzo2024vase}. Despite these advancements, explicit camera motion control remains a relatively underexplored area. Existing approaches often rely on predefined motion categories~\citep{guo2023animatediff,blattmann2023svd} or learnable LoRA modules~\citep{hu2021lora}. While methods like MotionCtrl~\citep{wang2023motionctrl} employ camera extrinsics, they exhibit limited precision in complex scenarios, and MultiDiff~\citep{muller2024multidiff} is constrained by class-specific training. More recently, several works~\citep{xu2024camco,he2024cameractrl,he2025cameractrlii} have leveraged Plücker coordinates~\citep{sitzmann2021light} for camera control, but still face challenges in producing realistic video outputs. Notably, the majority of current research generates videos as 2D frame sequences, largely overlooking the joint generation of dynamic 3D representations.

\vspace{-0.4cm}
\paragraph{Static 3D Scene Generation}
Recent progress in generative models~\citep{ho2020denoising,rombach2022high, yang2025fast3r, wang2025vggt} and 3D representations~\citep{kerbl20233dgs,mildenhall2020nerf} has significantly advanced static 3D scene generation. One prominent research direction focuses on structured scene generation from layouts or graphs~\citep{gao2023graphdreamer,bai2023componerf,po2023compositional,vilesov2023cg3d,yuan2025immersegen,lin2025partcrafter,lin2024instructscene,lin2024instructlayout}. Another line of research, more related to our work, addresses open-world scene generation from weak conditioning signals like text~\citep{chung2023luciddreamer,zhou2024dreamscene360} or images~\citep{chung2023luciddreamer, yu2024wonderworld,liang2024wonderland,huang2025voyager,li2025flashworld}. These methods often rely on image diffusion models~\citep{ho2020denoising,rombach2022high} as a backbone to provide strong 3D priors~\citep{chung2023luciddreamer,zhou2024dreamscene360,yu2024wonderworld,szymanowicz2025bolt3d,linDiffSplatRepurposingImage2025,wewer2024latentsplat}. The rise of video diffusion models has also motivated studies~\citep{liang2024wonderland,liu2024reconx, yu2024viewcrafter,sun2024dimensionx} to leverage them for improved 3D-aware consistency. Our work distinguishes itself by pioneering dynamic scene generation, addressing the critical challenge of modeling motion.

\vspace{-0.2cm}
\paragraph{Dynamic 3D Scene Generation}
Static 3D generation methods are inherently limited to motionless scenes. The natural, albeit challenging, progression is dynamic 4D scene generation~\citep{zhao2024genxd, zhang2024monst3r,chu2024dreamscene4d,liang2024feed,lin2025omniphysgs,li20244k4dgen,Video4DGen,team2025aether}. For example, Lyra~\citep{bahmani2025lyra} proposes a self-distillation framework that aims to distill the implicit 3D knowledge in the video diffusion models into an explicit 3DGS representation. Due to dataset limitations~\citep{RealEstate10k, dai2017scannet, yeshwanth2023scannet++, ling2024dl3dv, yu2023mvimgnet,huang2026thinking,wen2025dynamicverse}, prior works often tackle sub-problems. Some methods require a video and multi-view images of the first frame~\citep{4real,wang20244real,xie2024sv4d}. Others generate Dynamic 3D Gaussian Splatting from monocular video~\citep{chu2025dreamscene4d,wu2024cat4d,liang2024feed,nutshell} or rely on costly per-scene optimization~\citep{lei2024mosca, li2023dynibar,zhao2024pgdvs, som, wu20244d,sun2024splatter}. Recent feed-forward works generate dynamic pointmaps~\citep{team2025aether,4DNeX}, but this kind of representation struggles to achieve photorealism, resulting in renderings with holes and artifacts. In contrast, our work introduces a generalizable method that generates an explicit deformation Gaussian field from a single image, without per-scene optimization.

\vspace{-0.2cm}
\section{Methodology} \label{sec:method}
\begin{figure*}[t]
    \begin{center}
    \includegraphics[width=1 \textwidth]{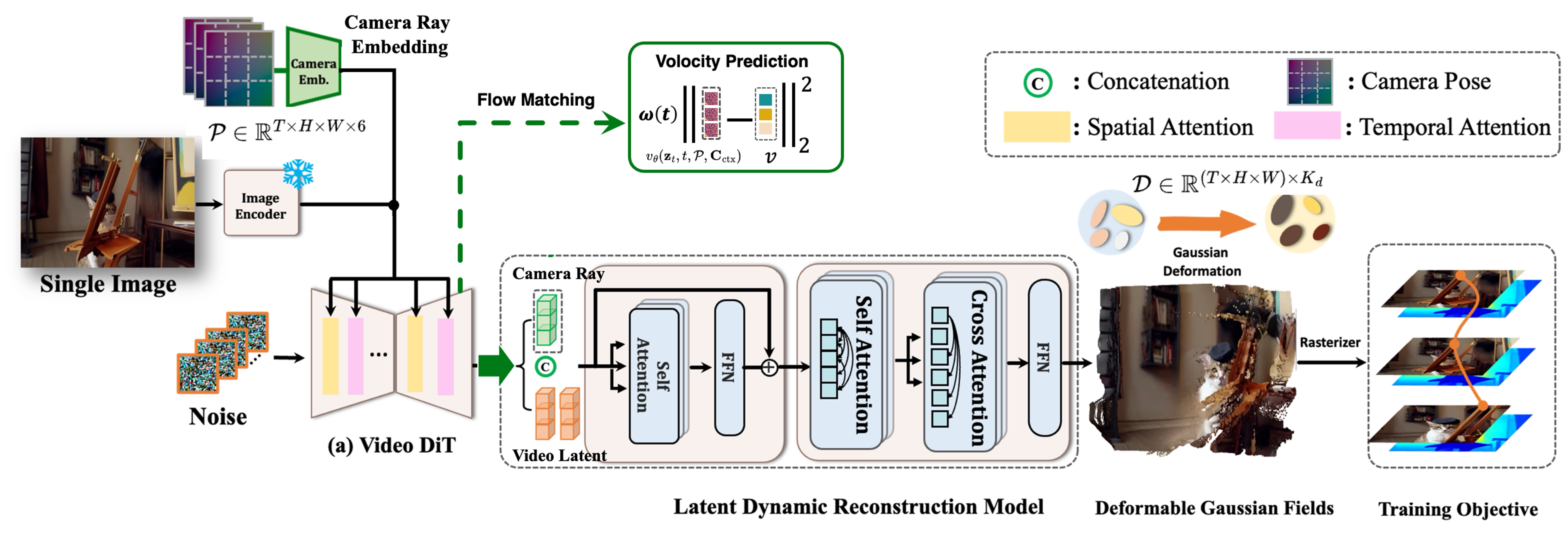}
    \end{center}
    \vspace{-0.2cm}
    \caption{\textbf{Architecture of \name{}.} We present a high-fidelity dynamic 3DGS generation method from a single image through four key innovations: (1) video diffusion latents processed by our novel Transformer (Sec.~\ref{LDRM}), (2) a dynamic 3DGS deformation mechanism (Sec.~\ref{Deformable Gaussian Fields}), (3) unified supervision with photometric, geometric, and motion losses (Sec.~\ref{Training Objective}), and (4) a progressive training scheme for robust geometry and texture.}
    \label{fig:architecture}
\end{figure*}

We address the generation of dynamic 4D scene representations from a single image $\mathbf{I}_0 \in \mathbb{R}^{H \times W \times 3}$, a text prompt $\mathbf{C}_{\text{ctx}}$, and camera poses $\mathcal{P} \in \mathbb{R}^{T \times H \times W \times 6}$ (Plücker embeddings~\citep{jia2020plucker}). Our methodology (Fig.~\ref{fig:architecture}) integrates a video diffusion model with a novel latent reconstruction Transformer, synergistically combining 2D appearance priors, geometry, and motion cues for high-fidelity 4D synthesis. First, a pre-trained video diffusion model, conditioned on $\mathbf{I}_0$ and $\mathcal{P}$, generates a latent tensor $\mathbf{z} \in \mathbb{R}^{n \times h \times w \times c}$, where $n$ is the feature count and $h, w, c$ are latent dimensions. Next, our Latent Dynamic Reconstruction Model (LDRM, Sec.~\ref{LDRM}) processes $\mathbf{z}$ and camera conditions to predict a deformable Gaussian field for novel view and time rendering. Second, we augment the static 3D Gaussian Splatting (3DGS)~\citep{kerbl20233dgs} with an efficient inter-frame deformation model (Sec.~\ref{Deformable Gaussian Fields}) to represent dynamics. Third, a unified supervision scheme (Sec.~\ref{Training Objective}) incorporates photometric, geometric, and motion losses. Finally, we employ a progressive training strategy to ensure high-fidelity textures and robust geometry.

\subsection{Data Curation} \label{data}
We start by developing a scalable 4D data annotation pipeline, meticulously designed to convert real-world videos into spatio-temporal point maps at metric scales. Our data curation strategy systematically integrates two complementary types of data sources:

(1) \textit{Synthetic Datasets}: We leverage seven synthetic datasets: TartanAir~\citep{wang2020tartanair}, MatrixCity~\citep{li2023matrixcity}, PointOdyssey~\citep{zheng2023pointodyssey}, DynamicReplica~\citep{karaev2023dynamicstereo}, Spring~\citep{mehl2023spring}, VKITTI2~\citep{cabon2020virtual}, and MultiCamVideo~\citep{bai2025recammaster}. These datasets provide precise ground-truth annotations for geometry and motion, with varied camera trajectories (both static and moving), which are essential for learning robust geometric and dynamic priors.
(2) \textit{Real-world Datasets}:  We incorporate two real-world datasets: RealEstate10K~\citep{zhou2018stereo} and Stereo4D~\citep{jin2024stereo4d}. These datasets offer authentic scene complexity and natural variations, which are crucial for enhancing the model's generalization capabilities. Since these datasets often feature stationary or limited camera motion and lack metric scale, we process them to serve as a basis for learning high-fidelity appearance. Inspired by~\citep{team2025aether}, we employ VideoDepthAnything~\citep{chen2025videodepthanything} and MegaSaM~\citep{li2025megasam} to recover metric scale from these datasets, enabling more precise camera control within our generative framework~\citep{bahmani2024ac3d}.
As shown in Algorithm~\ref{ALG:DepthAlignmentFinal}, through this comprehensive processing pipeline, we amass approximately 130,000 high-quality 4D training scenes. Following a rigorous quality control protocol, which includes dynamic object masking and reprojection error filtering.  We construct a collection of 130,000 diverse videos featuring dynamic scenes captured by stationary cameras. Real-world datasets such as RealEstate10K~\cite{RealEstate10k} only provide relative camera parameters estimated via COLMAP~\citep{schonberger2016structure}, resulting in an unknown global scale.
To address this, we re-estimate both metric depth maps and camera extrinsics using recent foundation models, Video Depth Anything~\citep{chen2025videodepthanything} and MegaSaM~\citep{li2025megasam}, to recover aligned geometry across frames.

\begin{algorithm}[!h]
    \caption{Metric Depth Reconstruction via Relative Depth Alignment}
    \label{ALG:DepthAlignmentFinal}
    \begin{algorithmic}[1]
        \State \textbf{Input:} RGB Image $I$, pre-trained DepthAnything model $\mathcal{F}_{DA}$, MegaSaM model $\mathcal{F}_{MS}$, metric depth oracle $\mathcal{P}_{M}$
        \State \textbf{Output:} Dense and metrically-scaled depth map $D^*$
        \Statex

        \State $D_{rel} \leftarrow \mathcal{F}_{DA}(I)$ \Comment{Generate relative depth map}
        \State $\mathcal{S} \leftarrow \mathcal{F}_{MS}(I)$ \Comment{Generate segmentation mask set}
        \State $\mathcal{A} \leftarrow \emptyset$ \Comment{Initialize anchor point set}
        \Statex

        \For{each mask $M_i \in \mathcal{S}$}
            \State $d_{gt, i} \leftarrow \mathcal{P}_{M}(M_i)$ \Comment{Query ground-truth metric depth for the mask}
            \If{$d_{gt, i}$ is a valid measurement}
                \State $V_i \leftarrow \{D_{rel}(u,v) \mid M_i(u,v) = 1\}$ \Comment{Extract corresponding relative depth values}
                \State $d_{rel, i} \leftarrow \text{median}(V_i)$ \Comment{Compute a robust representative value}
                \State $\mathcal{A} \leftarrow \mathcal{A} \cup \{(d_{rel, i}, d_{gt, i})\}$ \Comment{Add the pair to the anchor set}
            \EndIf
        \EndFor
        \Statex

        \State \Comment{Estimate optimal scale and shift by solving the least-squares problem}
        \State $(s^*, t^*) \leftarrow \underset{s, t}{\arg\min} \sum_{(d_{rel, i}, d_{gt, i}) \in \mathcal{A}} (s \cdot d_{rel, i} + t - d_{gt, i})^2$
        \Statex

        \State \Comment{Apply the transformation to the full relative depth map}
        \State $D^* \leftarrow s^* \cdot D_{rel} + t^*$
        \Statex
        
        \State \Return $D^*$
    \end{algorithmic}
\end{algorithm}

\subsection{Latent Dynamic Reconstruction Model} \label{LDRM}

Directly applying video diffusion models for 3D-aware latent synthesis is challenging due to their lack of explicit camera control and inconsistent dynamic content, which hinders robust 3D reconstruction. Inspired by recent latent diffusion models~\citep{blattmann2023align, latentdiffusion, humansplat, liangWonderlandNavigating3D2024}, we propose the \textbf{L}atent \textbf{D}ynamic \textbf{R}econstruction \textbf{M}odel (\textbf{LDRM}) to avoid costly per-scene optimization. LDRM employs a pre-trained video diffusion model, conditioned on an input image and camera poses, to generate a compact, 3D-aware latent tensor $\mathbf{z}$ that ensures structural and appearance consistency across views, making it ideal for 3D lifting. Given the latent tensor $\mathbf{z} \in \mathbb{R}^{n \times h \times w \times c}$ and camera poses, we create and concatenate latent and pose tokens of identical length. These are processed by Transformer blocks~\citep{GQA:2023}, after which a lightweight decoder regresses 3D Gaussian attributes. A 3D deconvolutional layer then maps these attributes to the source video pixels.


\subsection{Deformable Gaussian Fields} \label{Deformable Gaussian Fields}
A static 3D scene can be represented as a collection of $\mathbf{M}$ Gaussian primitives $\{ \boldsymbol{G}_p \}_{p=1}^\mathbf{M}$. Each Gaussian $\boldsymbol{G}_p$ is characterized by its mean location $\boldsymbol{\mu}_p \in \mathbb{R}^3$, scaling factors $\boldsymbol{s}_p \in \mathbb{R}^3$, orientation quaternion $\boldsymbol{q}_p \in \mathbb{R}^4$, opacity $\alpha_p \in \mathbb{R}$, and color features $\boldsymbol{c}_p \in \mathbb{R}^C$. We use Spherical Harmonics~(SH) to model view-dependent effects. The spatial influence of each Gaussian is given by:
\begin{equation}
 \boldsymbol{G}_p(\mathbf{x}) := \exp \left( -\frac{1}{2} (\mathbf{x}-\boldsymbol{\mu}_p)^\top \boldsymbol{\Sigma}_p^{-1} (\mathbf{x}-\boldsymbol{\mu}_p) \right),
\end{equation}
where $\boldsymbol{\Sigma}_p$ is the covariance matrix derived from $\boldsymbol{s}_p$ and $\boldsymbol{q}_p$. Inspired by ~\citep{yang2024deformable,Gaussian-flow,Gaufre}, we introduce a deformable 3D Gaussian formulation to represent dynamic scene. For each Gaussian $p$ at time step $t$, the predicted deformation field comprises a displacement for its mean, $\Delta\boldsymbol{\mu}_p^t \in \mathbb{R}^3$; an adjustment to its rotation, $\Delta\boldsymbol{q}_p^t \in \mathbb{R}^4$; and a modification to its scale, $\Delta\boldsymbol{s}_p^t \in \mathbb{R}^3$. The deformed parameters at time $t$ are updated as follows: $\boldsymbol{\mu}_p^t := \boldsymbol{\mu}_p^0 + \Delta\boldsymbol{\mu}_p^t$, $\boldsymbol{q}_p^t := \boldsymbol{q}_p^0 \otimes \Delta\boldsymbol{q}_p^t$ (quaternion multiplication), and $\boldsymbol{s}_p^t := \boldsymbol{s}_p^0 + \Delta\boldsymbol{s}_p^t$. These deformed Gaussians are then rendered using a differentiable Gaussian rasterization pipeline. Deformable Gaussian Fields is equipped with the LDRM, which generates a Gaussian feature map $\boldsymbol{G} \in \mathbb{R}^{(T \times H \times W) \times K_g}$, where $K_g$ denotes the number of parameters for each Gaussian primitive. Concurrently, the LDRM predicts a corresponding deformation map $\mathcal{D} \in \mathbb{R}^{(T \times H \times W) \times K_d}$. The dimensionality of this deformation, $K_d=10$, comprises offsets for the mean ($\Delta\boldsymbol{\mu} \in \mathbb{R}^3$), rotation ($\Delta\boldsymbol{q} \in \mathbb{R}^4$), and scale ($\Delta\boldsymbol{s} \in \mathbb{R}^3$).

 During training and inference, we employ a pruning strategy based on the opacity values predicted by the model, removing Gaussians with opacity below a threshold $\tau_{opacity}=0.005$ for rendering.

\subsection{Training Objective} \label{Training Objective}

To enhance the geometric consistency of the generated latents, we introduce a training scheme that jointly optimizes the network across multi-tasks via differentiable rendering. The total loss is a weighted sum of four components:
\begin{equation}
    \mathcal{L} = \mathcal{L}_{\text{FM}} + \lambda_{photo}\mathcal{L}_{\text{photo}} + \lambda_{geo}\mathcal{L}_{\text{geo}} + \lambda_{motion}\mathcal{L}_{\text{motion}},
\end{equation}
where $\lambda_{photo}=1.0$, $\lambda_{geo}=0.5$, and $\lambda_{motion}=2.0$ are balancing coefficients.


\vspace{-0.4cm}
\paragraph{Flow Matching Loss}
The Flow Matching (FM)~\citep{lipman2023flow} approach learns the vector field that transports a noise distribution to the data distribution. The FM loss, $\mathcal{L}_{\text{FM}}$, is applied only to the parameters of the base video diffusion model. It is used to finetune the pretrained video model on our curated dataset to better align its latent space with 4D-consistent data. The LDRM and Gaussian prediction head are trained exclusively with the rendering-based losses described below. Let $\mathbf{z}^{(0)}$ be a clean latent sequence from the data distribution $p_{\text{data}}$, and $\mathbf{z}^{(1)} \sim \mathcal{N}(0, \mathbf{I})$ be a sample from the prior gaussian noise. 
The FM objective is to train a vector field model $v_\theta(\mathbf{z}^{(t)}, t)$ to match the ground-truth vector field $u_t(\mathbf{z}^{(t)})$.
\begin{equation}
    \mathcal{L}_{\text{FM}}(\theta) = \mathbb{E}_{t, p_t(\mathbf{z}^{(t)})} \left[ \| v_\theta(\mathbf{z}^{(t)}, t) - u_t(\mathbf{z}^{(t)}) \|_2^2 \right].
\end{equation}
This loss encourages the diffusion model to generate latent videos that are faithful to the distribution of our 4D-annotated dataset.

\vspace{-0.4cm}
\paragraph{Photometric Loss}
To facilitate high-quality novel view synthesis, we optimize the 3DGS parameters using a composite loss:
\begin{align}
    \mathcal{L}_{\text{photo}} = \texttt{MSE}(\hat{\mathbf{I}}^k, \mathbf{I}^k) + \lambda_p \cdot \texttt{LPIPS}(\hat{\mathbf{I}}^k, \mathbf{I}^k),
\end{align}
where $\hat{\mathbf{I}}^k$ is the rendered image for view $k$, $\mathbf{I}^k$ is the ground-truth image, and $\lambda_p$ is a balancing coefficient for the LPIPS~\citep{zhang2018unreasonable} term.

\vspace{-0.1cm}
\paragraph{Geometric Loss}
Inspired by ~\citep{4K4DGen}, we introduce a geometric regularization term to enforce accurate depth relationships. Let $\hat{D}_k$ be the rendered depth map and $D_k^*$ be the ground-truth depth for view $k$. 
$
    \mathcal{L}_{\text{geo}}(\hat{D}_k, D_k^*) = 1 - \frac{\texttt{Cov}(\hat{D}_k, D_k^*)}{\sqrt{\texttt{Var}(\hat{D}_k)\texttt{Var}(D_k^*)}},
$
where $\texttt{Cov}$ and $\texttt{Var}$ are covariance and variance functions. We also apply a total variation loss, $\mathcal{L}_{\text{TV}} = \|\nabla \hat{D}_k\|_1$, to enforce local smoothness.

\vspace{-0.1cm}
\paragraph{Motion Loss}
Given 3D point tracking data, the ground-truth motion for a point $j$ is its displacement $\Delta\mathbf{x}_j$. For synthetic data, these tracks are readily available. For real-world videos, we utilize the CoTracker~\citep{karaev2024cotracker} model to obtain point tracks, which are then filtered based on confidence scores to ensure high quality. The motion loss is:
\begin{equation}
    \mathcal{L}_{\text{motion}} = \frac{1}{|\mathcal{O}|} \sum_{j \in \mathcal{O}} \left( \lambda_{m} \| \Delta\hat{\mathbf{x}}_j - \Delta\mathbf{x}_j \|_2 +  \|\Delta\hat{\mathbf{x}}_j\|_1 \right),
\end{equation}
where $\mathcal{O}$ is the set of tracked points, $\Delta\hat{\mathbf{x}}_j$ is the predicted displacement, and $\lambda_{m}$ is a weighting coefficient. 

\vspace{-0.1cm}
\paragraph{Progressive Training Scheme}\label{sec:3.4}

\ding{202} \textbf{Static Geometry Pre-training (40K iterations).} We first establish a strong geometric prior by training LDRM on static scenes (e.g., TartanAir, RealEstate10K) at a low resolution (256 $\times$ 256), using only photometric and geometric losses. During this stage, the deformation module (an 8-layer DPT head) is frozen.

\ding{203} \textbf{High-Resolution Refinement (40K iterations).} With the deformation module still frozen, we enhance reconstruction fidelity by training on static scenes under a high resolution (512 $\times$ 512).

\ding{204} \textbf{Dynamic Scene Fine-tuning (20K iterations).} Finally, we unfreeze and fine-tune \textit{the entire model} on dynamic datasets (PointOdyssey, DynamicReplica, Spring, VKITTI2, and Stereo4D). This stage employs the complete loss function, including a motion loss term, to learn temporal deformations.
This progressive strategy, combined with our large-scale 4D dataset, enables our model to learn complex dynamics and generate high-fidelity, temporally coherent dynamic scenes.



\vspace{-0.2cm}
\section{Experimental Evaluation}\label{sec:exp}

\begin{table*}[!th] 
    \centering 
    \caption{Quantitative comparison of appearance fidelity and aesthetic quality. ${\dagger}$ denotes methods requiring per-scene optimization. We highlight \textbf{first-place} and \underline{second-place} results.} 
    \label{tab:quantitative_video_aesthetic}
    \renewcommand\arraystretch{1.2}
    \resizebox{0.9\textwidth}{!}{%
    \begin{tabular}{@{}lrrrrrc@{}}
    \toprule[1.2pt] 
    \multirow{2}{*}{\makecell[c]{Method}} & \multicolumn{5}{c}{\textbf{Video Generation \& Aesthetic Quality}} & \multirow{2}{*}{\makecell[c]{Rec. Time \\ (lower is better)}} \\ 
    \cmidrule(lr){2-6} 
    & FVD $\downarrow$ & KVD $\downarrow$ & CLIP-Score $\uparrow$ & CLIP-Aesthetic $\uparrow$ & QA-Quality $\uparrow$ & \\ 
    \midrule
    \multicolumn{7}{l}{\textit{Camera-Controlled Video Generation}} \\
    \midrule
    CameraCtrl~\citep{he2024cameractrl} & 478.192 & 8.105 & 19.365 & 2.965 & 1.894 & 20s \\ 
    AC3D~\citep{bahmani2024ac3d} & 339.431 & 6.342 & 20.673 & 3.324 & 2.158 & 28s \\
    \midrule
    \multicolumn{7}{l}{\textit{Explicit 3DGS Representation (Optimization-based)}} \\
    \midrule
    AC3D + Shape of Motion$^{\dagger}$~\citep{som} & 373.045 & 6.511 & 16.201 & 3.043 & 1.838 & \underline{18min} \\ 
    AC3D + SaV$^{\dagger}$~\citep{sav} & 327.122 & 5.816 & 19.018 & 4.371 & 2.382 & 35min \\ 
    AC3D + Mosca$^{\dagger}$~\citep{lei2024mosca} & \underline{235.961} & \textbf{2.012} & \underline{20.214} & \underline{4.999} & \textbf{2.842} & 45min \\ 
    \midrule 
    \rowcolor{mygray}
    \textbf{Ours (Feed-forward)} & \textbf{210.153} & \underline{2.316} & \textbf{23.123} & \textbf{5.231} & \underline{2.813} & \textbf{30s} \\ 
    \bottomrule[1.2pt] 
    \end{tabular}%
    } 
\end{table*}

\begin{figure*}[!th] 
  \centering 
  \includegraphics[width=\linewidth]{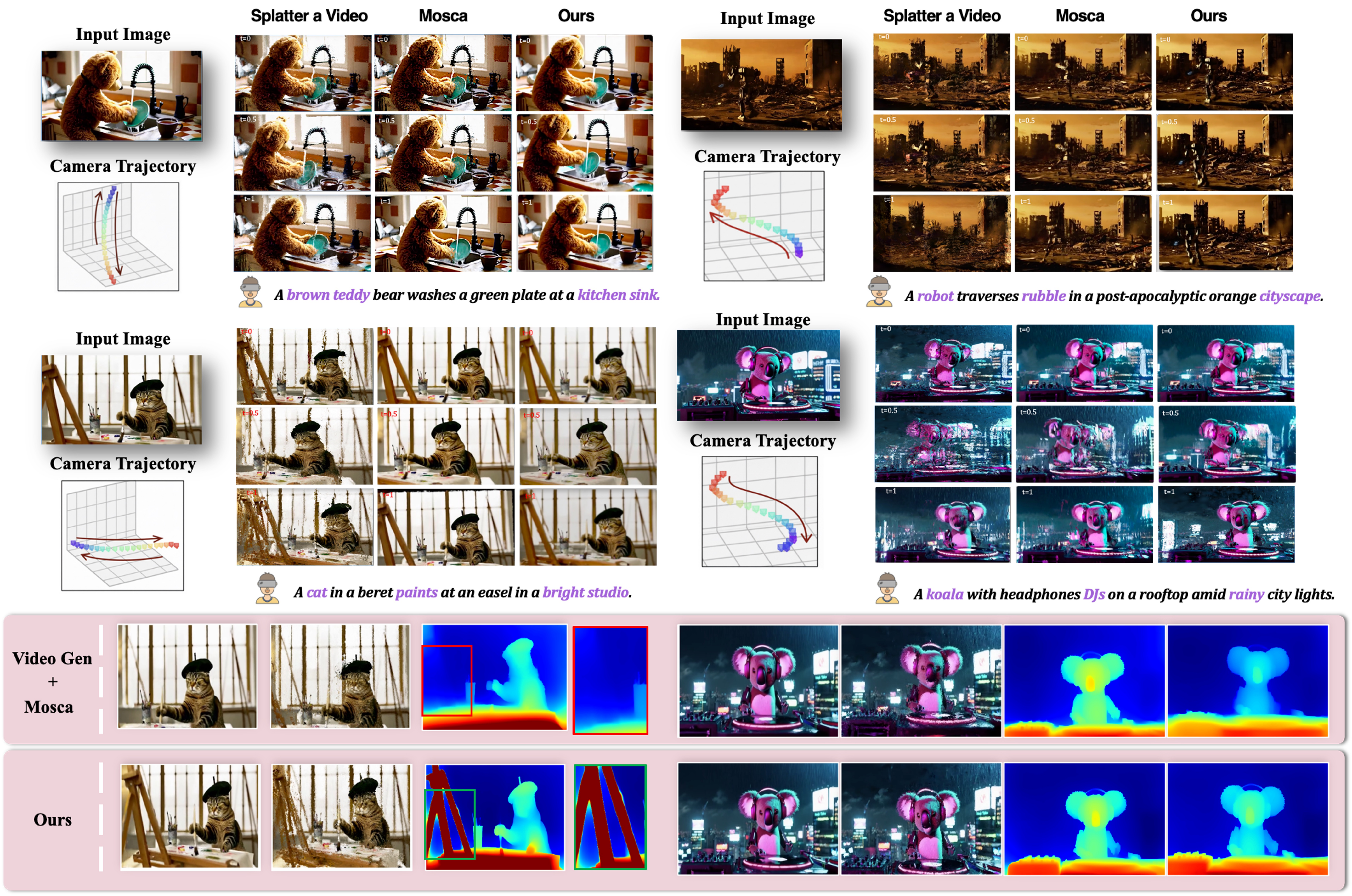} 
  \caption{\textbf{Qualitative comparison with state-of-the-art methods.} \name{} (last column) generates more visually appealing and temporally consistent 4D scenes with superior geometric fidelity compared to baselines. Kindly zoom in for details.}
  \label{fig:res1} 
\end{figure*}

\subsection{Implementation Details} \label{subsec:Implementation}
Our framework is built upon CogVideoX~\citep{yang2024cogvideox}, a pretrained Video Diffusion Transformer model operating in the latent space of a 3D Causal Variational Autoencoder with a $32 \times 4 \times 8 \times 8$ compression scheme. The architecture consists of 32 blocks and a hidden dimensionality of 4096. The Latent Diffusion model for Deformable Radiance Fields (LDRM) comprises 16 standard Transformer blocks. Latent features, with a channel dimension of $c=32$, are projected into a 64-dimensional embedding space before being processed by the Transformer backbone. The latent tensor is tokenized with a patch size of $2 \times 2$, resulting in a sequence of tokens for the transformer. To incorporate textual guidance, each DiT block is equipped with a cross-attention layer that integrates image embeddings from a T5 model~\citep{T5}.

For training, we utilize the AdamW optimizer~\citep{adamw} with an initial learning rate of $10^{-5}$ and a weight decay of $10^{-4}$. The loss weighting hyperparameters are set to $\lambda_{photo}=1.0$, $\lambda_{geo}=0.5$, and $\lambda_p=0.5$ for the photometric loss and $\lambda_m=2$ for the motion loss. The model is trained for 100,000 iterations using a cosine learning rate scheduler, requiring approximately 7 days on 32 A100 GPUs with BF16 mixed precision. At inference, our model generates a dynamic scene in approximately \textbf{30 seconds}.

\subsection{Evaluation Protocol} \label{sec:evaluation}
\paragraph{Baselines}
We benchmark our single-stage pipeline against two-stage approaches that combine state-of-the-art techniques. Specifically, we employ AC3D~\citep{bahmani2024ac3d} for single-image controllable video generation, followed by Mosca~\citep{lei2024mosca} for dynamic 3D Gaussian reconstruction. For a comprehensive evaluation of camera controllability, we created an evaluation set of \textbf{160 samples} by applying five distinct camera trajectories (spiral, forward, backward, upward, and downward) to 32 unique text-captioned scenes. We also compare against recent feed-forward 4D generation methods where possible, though many are not open-sourced or have different input requirements, precluding a direct, fair comparison on our benchmark.

\paragraph{Metrics}
Our evaluation assesses both prompt-scene consistency and aesthetic quality using several established metrics: CLIP similarity score~\citep{radford2021learning}, CLIP-Aesthetic score~\citep{aesthetic_predictor}, and the VLM-based visual scorer Q-Align (QA-Quality)~\citep{wu2023qalign}. Video quality is measured using Fréchet Video Distance (FVD)~\citep{unterthiner2019fvd} and Kernel Video Distance (KVD)~\citep{unterthiner2018towards}. To evaluate geometric integrity, we use MASt3R~\citep{mast3r_arxiv24} for local correspondence matching between input and generated novel views, reporting the average number of matches, as well as subject and background consistency scores~\citep{zheng2025vbench2}.

\subsection{Quantitative and Qualitative Evaluation} \label{subsec:qualitative_evaluation}

\begin{table}[t]
    \centering
    \caption{Geometric integrity and reconstruction time. ${\dagger}$ denotes optimization-based methods. Best results are in \textbf{bold}, second best are \underline{underlined}.}
    \label{tab:quantitative_geometric_reconstruction}
    \renewcommand\arraystretch{1.1}
    \resizebox{\columnwidth}{!}{%
    \begin{tabular}{@{}lcccc@{}} 
        \toprule[1.2pt]
        Method & \makecell{Avg. \\ Matches $\uparrow$} & \makecell{Subj. Consist. \\ Score $\uparrow$} & \makecell{Bg. Consist. \\ Score $\uparrow$} & \makecell{Time \\ $\downarrow$} \\ 
        \midrule
        \multicolumn{5}{@{}l}{\textit{Camera-Controlled Video Generation}} \\
        \midrule
        CameraCtrl~\citep{he2024cameractrl} & 2015.82 & 72.25 & 74.53 & 20s \\ 
        AC3D~\citep{bahmani2024ac3d} & 2489.16 &  75.64  & 75.91 & 28s \\   
        \midrule
        \multicolumn{5}{@{}l}{\textit{Explicit 3DGS Representation}} \\
        \midrule
        AC3D + Shape of Motion$^{\dagger}$~\citep{som} & 2874.22 & 83.13  & 83.33  & 18min \\   
        AC3D + SaV$^{\dagger}$~\citep{sav} & 3035.43 & 85.96 & 84.23 & 35min \\
        AC3D + Mosca$^{\dagger}$~\citep{lei2024mosca} & \underline{4500.68} & \underline{86.23} & \textbf{90.43} & 45min \\ 
        \midrule
        \rowcolor{mygray}
        \textbf{Ours} & \textbf{5114.22} & \textbf{88.32} & \underline{89.89} & \textbf{30s} \\ 
        \bottomrule[1.2pt]
    \end{tabular}%
    }
\end{table}

\begin{table}[t]
    \centering
    \caption{Comparison of Average Relative Pose Error (RPE), highlighting our explicit model's superior accuracy in translation and rotation, alongside its additional capabilities.}
    \label{tab:rpe_comparison_scaled}
    \renewcommand\arraystretch{1.2}
    \resizebox{\columnwidth}{!}{%
    \begin{tabular}{l|ccccc}
        \toprule[1.2pt]
        \textbf{Method} & \makecell{Avg. RPE \\ (Translation) $\downarrow$} & \makecell{Avg. RPE \\ (Rotation) $\downarrow$} & \makecell{Novel View \\ Synthesis} & \makecell{Depth \\ Rasterization} & \makecell{Real-time \\ Interaction} \\
        \midrule
        AC3D~\citep{bahmani2024ac3d} & 3.001 & 0.810 & \ye & \no & \no \\
        \rowcolor{mygray}
         \textbf{Ours (Explicit)} & \textbf{0.012} & \textbf{0.008} & \ye & \ye & \ye \\
        \bottomrule[1.2pt]
    \end{tabular}%
    }
\end{table}

\begin{figure}[h]
\begin{center}
\includegraphics[width=\textwidth]{suppx.pdf}
\end{center}
\caption{\textbf{Qualitative results under extreme viewpoints.}}
\label{fig:more_res}
\end{figure}


\paragraph{Quantitative Results}
As detailed in Tab.~\ref{tab:quantitative_video_aesthetic} and Tab.~\ref{tab:quantitative_geometric_reconstruction}, our approach demonstrates competitive or superior performance across a range of metrics. For video generation and aesthetic quality (Tab.~\ref{tab:quantitative_video_aesthetic}), our method is highly competitive, while drastically reducing the reconstruction time to just 30 seconds. This represents a substantial efficiency gain over optimization-based methods like ``AC3D + Mosca,'' which requires 45 minutes. Furthermore, Tab.~\ref{tab:quantitative_geometric_reconstruction} shows that our method achieves precise, camera-controllable generation with strong geometric integrity. 
\vspace{-0.2cm}
\paragraph{Qualitative Results} 
As illustrated in Fig.~\ref{fig:res1}, our method generates 4D scenes that are more visually appealing, temporally coherent, and geometrically accurate than baseline methods. For instance, our generated videos exhibit smoother motion and fewer artifacts compared to SaV and Mosca. This visual superiority is a direct result of our model's ability to predict a deformable 3D Gaussian field, which provides a continuous and comprehensive representation of the scene's temporal evolution. In contrast, the dynamic motion capabilities of camera control models like AC3D and CameraCtrl are constrained by their 2D video priors, often leading to less dynamic results.
\vspace{-0.2cm}
\paragraph{Generation Controllability} 
A key advantage of our explicit 4D representation is the ability to enforce physical consistency by rendering video deterministically from a specified camera path. We validate this by quantifying camera pose fidelity using the Relative Pose Error (RPE) metric~\citep{sturm2012benchmark}. As shown in Table~\ref{tab:rpe_comparison_scaled}, our method significantly improves pose accuracy compared to the implicit baseline.
\vspace{-0.2cm}
\subsection{Ablation and Analysis}\label{subsec:ablation}

\begin{table}[t] 
    \centering 
    \caption{\textbf{Ablation Study on Motion Loss.} We evaluate the impact of our proposed motion loss on dynamic video generation.} 
    \label{tab:ablation_motion_loss} 
    \renewcommand\arraystretch{1.2} 
    \resizebox{\columnwidth}{!}{%
    \begin{tabular}{l|ccccccc} 
        \toprule[1.2pt] 
        \textbf{Method} & \makecell{FVD  $\downarrow$} & \makecell{KVD  $\downarrow$} & \makecell{QA-Quality  $\uparrow$} & \makecell{Avg. Matches $\uparrow$} & \makecell{Subj. Consist. \\ Score $\uparrow$} & \makecell{Bg. Consist. \\ Score $\uparrow$} & \makecell{Rec. Time  $\downarrow$} \\ 
        \midrule 
        w/o motion loss & 351.382 & 3.351 & 2.145 & 4821.56 & 82.45 & 85.12 & 30s \\ 
        \rowcolor{mygray} 
        \textbf{Ours} & \textbf{210.153} & \textbf{2.316} & \textbf{2.813} & \textbf{5114.22} & \textbf{88.32} & \textbf{89.89} & \textbf{30s} \\ 
        \bottomrule[1.2pt] 
    \end{tabular}%
    }
\end{table}

\begin{figure*}[h!]
    \centering
    \includegraphics[width=\linewidth]{figures/diff_application.pdf}
    \caption{Applications of \name{}. Our method supports applications such as novel view synthesis and depth map extraction from the generated 4D representation.}
    \label{fig:application}
\end{figure*}

\begin{figure}[t]
    \centering
    \includegraphics[width=\linewidth]{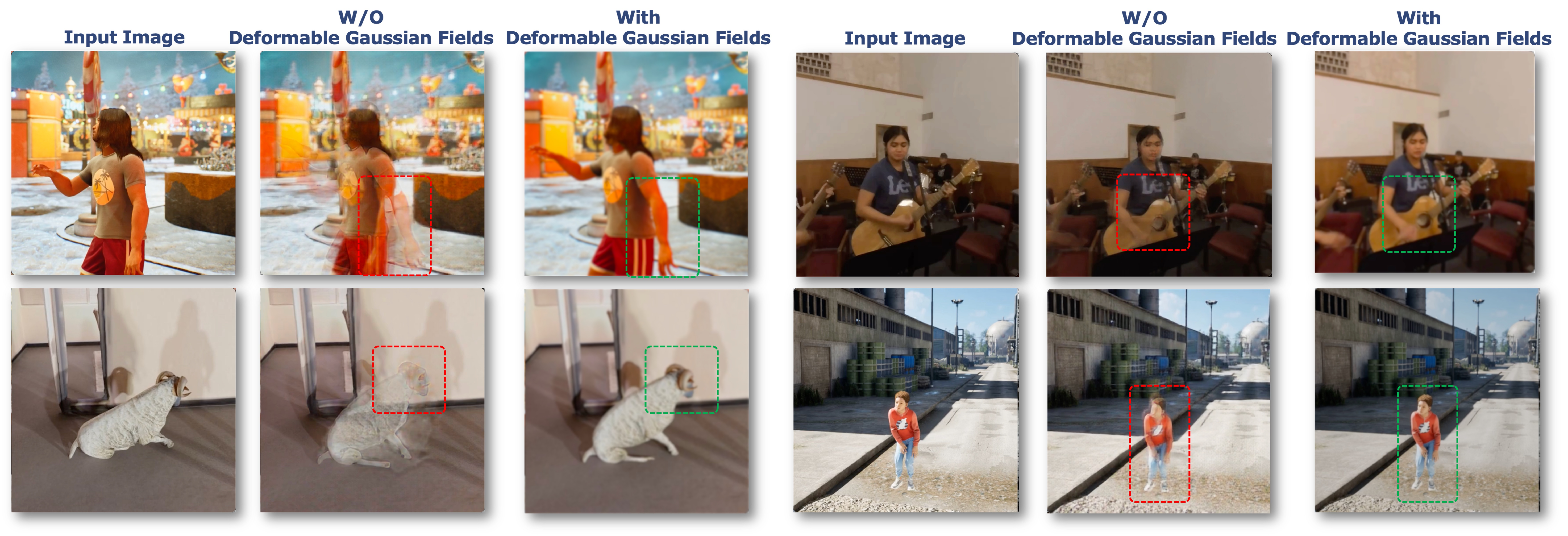} 
    \caption{Ablation of the \textbf{Deformation Gaussian Field}. Removing this module (the \textcolor{red}{red} bounding boxes) results in ghosting artifacts, particularly in frames with large motion.}
    \label{fig:deformation}
\end{figure}

\begin{figure}[t]
    \centering
    \includegraphics[width=\linewidth]{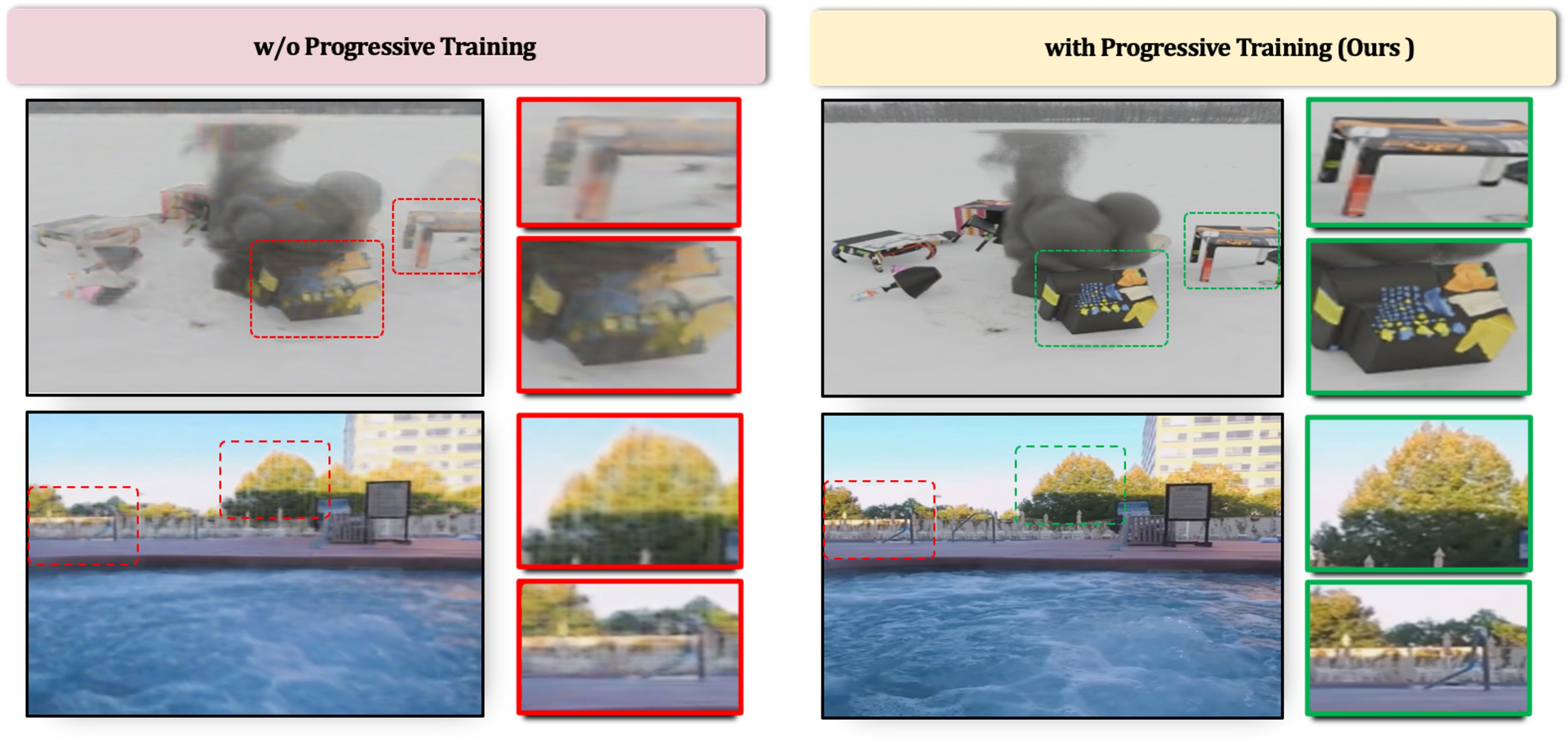} 
    \caption{Ablation on the progressive training strategy. Our approach (left) yields higher visual quality than direct dynamic training (right) after 100K iterations.}
    \label{fig:train}
\end{figure}



\paragraph{Effect of Deformation Gaussian Field}
Fig.~\ref{fig:deformation} illustrates the importance of the deformation Gaussian module. Without this module, the model struggles to differentiate between camera movement and the motion of foreground objects. This inability to properly combine 3D Gaussian splats from different timestamps leads to motion blur, spike artifacts, and a general degradation in image quality. By employing the deformation Gaussian field, our model effectively fuses reconstruction information from various moments, thereby achieving higher visual quality.

\vspace{-0.2cm}
\paragraph{Effect of Explicit Representation}
Our adoption of an explicit 3D Gaussian Splatting representation offers several key advantages over implicit models, as detailed in Table~\ref{tab:rpe_comparison_scaled}. Firstly, it enables superior camera controllability, drastically reducing the Relative Pose Error (RPE) in both translation and rotation. This ensures that the generated video faithfully adheres to the specified camera path. Secondly, the explicit nature of the representation unlocks additional functionalities not available in the implicit baseline, such as depth rasterization and real-time interaction. 

\paragraph{Effect of Motion Loss}
We perform an ablation study on our proposed deformation module and motion loss. As shown in Table~\ref{tab:ablation_motion_loss}, removing this component prevents the network from accurately modeling temporal deformations, which is critical for dynamic video synthesis. Its absence significantly degrades scene reconstruction quality and negatively impacts all quantitative metrics.

\vspace{-0.2cm}
\paragraph{Importance of Progressive Training} 
Our progressive training strategy is crucial for robust 4D scene generation. Directly training on dynamics without static pre-training fails to initialize the 3DGS, leading to unstable training and degraded 4D scene quality. In contrast, our approach first establishes a static scene representation, providing a solid foundation for learning complex dynamics, which significantly enhances performance and visual fidelity. Direct dynamic training is inefficient, either converging to a suboptimal state or requiring triple the training time (e.g., 21 versus 7 days) for a comparable baseline. As shown in Fig.~\ref{fig:train}, our progressive strategy yields substantially higher visual quality after 100K iterations, underscoring its superior resource efficiency and final performance.
        
\vspace{-0.2cm}
\paragraph{Applications}
 Beyond controllable video generation, our framework naturally supports novel view synthesis for dynamic scenes, enabling free-viewpoint rendering. The explicit 3DGS representation facilitates real-time rendering and interaction, paving the way for immersive virtual reality experiences. 
Furthermore, geometric information, such as depth maps, can be readily extracted. We showcase additional in-the-wild cases in Fig.~\ref{fig:more_res}, followed by comprehensive qualitative results for these applications in Fig.~\ref{fig:application}.

\section{Conclusion}\label{sec:conclu}
In this work, we present a novel framework for explicit deformation Gaussian field generation from a single image in a \textit{feed-forward} manner and achieves three key innovations: (1) unified diffusion transformer architecture integrating dynamic scene modeling, (2) geometry-aware latent representation enabling efficient view synthesis, (3) real-time rendering pipeline supporting practical applications.

\clearpage

{
\small
\bibliographystyle{ieeenat_fullname}
 \bibliography{references}
}

\appendix
\maketitlesupplementary
\setcounter{figure}{0}
\setcounter{table}{0}
\renewcommand{\thefigure}{\Alph{figure}}
\renewcommand{\thetable}{\Alph{table}}





\begin{table*}[h]
    \centering
    \caption{\textbf{Training Datasets Statistics.} Overview of the datasets used for training \name~at scale, highlighting their dynamic nature, multi-camera setups, depth annotations, tracking capabilities, and real-world applicability.
   }
    \label{tab:datasets}
    \renewcommand\arraystretch{1}
    \resizebox{\textwidth}{!}{

    \begin{tabular}{l|cccccrr}
        \toprule[1.2pt]

        \textbf{Dataset} & \highlightcell{Dynamic?} & \highlightcell{Multi-camera?} & \highlightcell{Depth?} & \highlightcell{Tracking?} & \highlightcell{Real?} & \highlightcell{\#Scenes} & \highlightcell{\#Frames}  \\
        \midrule
        TartanAir~\citep{wang2020tartanair} & \no & \no & \ye & \no & \no & 0.4K & 0.49M \\
        MatrixCity~\citep{li2023matrixcity} & \no & \no & \ye & \no & \no & 4.5K & 0.31M  \\

        RealEstate10K~\citep{zhou2018stereo} & \no & \no & \no & \no & \ye & 70K & 6.36M  \\
        PointOdyssey~\citep{zheng2023pointodyssey} & \ye & \no & \ye & \ye & \no & 0.1K & 0.18M  \\
        DynamicReplica~\citep{karaev2023dynamicstereo} & \ye & \no & \ye & \ye & \no & 0.5K & 0.26M  \\
        Spring~\citep{mehl2023spring} & \ye & \no & \ye & \no & \no & 0.03K & 0.003M  \\
        VKITTI2~\citep{cabon2020virtual} & \ye & \no & \ye & \no & \no & 0.1K & 0.03M  \\
        MultiCamVideo~\citep{bai2025recammaster} & \ye & \ye & \no & \no & \no & 14K & 11M \\
        
        Stereo4D~\citep{jin2024stereo4d} & \ye & \no & \ye & \ye & \ye & 74K & 14.8M \\
        \bottomrule[1.2pt]
    \end{tabular}
    }
    \vspace{-0.2cm}
\end{table*}
\section{Dataset Curation Details} \label{apx:dataset}
We construct a collection of 130,000 diverse videos. A key challenge is obtaining metric scale for real-world datasets like RealEstate10K, which only provide relative camera poses from COLMAP. Our metric scale estimation procedure, addresses this by anchoring the relative depth predictions from foundation models to a sparse set of metric-scale 3D points. These anchor points are obtained by running Structure-from-Motion (SfM) and then manually scaling the reconstruction using known real-world information, such as the height of the camera or the size of objects in the scene for a small subset of scenes. This provides a robust mechanism for recovering metric scale, which is crucial for training a model with precise camera control. The error of this estimation is typically low, with the scale factor showing a variance of less than 5\% across different subsets of anchor points.

\section{More Implementation Settings}  \label{apx:results}

\paragraph{Reproducibility} 
To facilitate reproducibility, we present our detailed experimental settings and evaluation metrics in Section ~\ref{apx:imple}. This section provides a comprehensive description of our implementation details. Moreover, \textbf{our source code, pre-trained models, and curated dataset will be publicly available upon publication.}

\subsection{Video Transformer Denosing Details} \label{apx:imple}

\paragraph{Details of Model Inputs}

The model is conditioned on a single source image and a predefined camera motion trajectory, such as spiral, forward, backward, upward, or downward. Accompanying this, a textual prompt is provided, which can either be automatically generated from the source image using a Multimodal Large Language Model (MLLM)~\citep{Qwen-VL} or set to a generic high-fidelity description, for instance, ``a scene with 4K ultra HD, surround motion, realistic tone, panoramic shot, wide-angle view, cinematic quality''.

\paragraph{Classifier-Free Guidance}  \textbf{C}lassifier-\textbf{F}ree \textbf{G}uidance (\textbf{CFG}) has emerged as a prevalent technique for balancing controllability and sample diversity in diffusion models. However, we observe that its uniform scaling mechanism inadvertently introduces ``over-sharpening artifacts'' in the final frames of generated orbital sequences. To mitigate this limitation, we introduce a cosine-based dynamic guidance schedule during the sampling of validation videos, formulated as:

\begin{equation}
\gamma(t) = 1 + \gamma_{\text{max}} \cdot \left( \frac{1 - \cos\left( \pi \left( \frac{N - t}{N} \right)^5 \right)}{2} \right)
\end{equation}

\noindent where $\gamma_{\text{max}}$ denotes the maximum guidance scale, $N$ represents the total number of inference steps, and $t$ is the current timestep. This adaptive scheduling progressively reduces guidance intensity in later denoising stages, effectively preserving temporal consistency while maintaining sample fidelity. In our experiments, we set the total number of inference steps $N=30$ and the maximum guidance scale $\gamma_{\text{max}}=7.5$.



 


\subsection{Deformation Field Generation} 

To predict the per-Gaussian deformations, our \textbf{LDRM} employs a lightweight spatio-temporal network. The network takes as input a latent representation of the scene at a canonical time step, conditioned on a time embedding for the target frame $t$. The architecture extracts features at multiple spatial resolutions to effectively capture both local and global motion patterns. The final layer of the network is a convolutional layer with a kernel size of $1 \times 1$, which projects the high-dimensional features into the final deformation map $\mathcal{D}$. This map has a dimensionality of $K_d=10$ channels, which directly correspond to the predicted mean displacement (3 channels), rotational delta quaternion (4 channels), and scaling adjustment (3 channels) for each Gaussian primitive. No activation function is applied to the output layers for displacement and scale, allowing for unbounded predictions. The output quaternion components are normalized to ensure they represent a valid rotation.
 
%



\subsection{Details of Progressive Training Scheme.}
Our progressive training scheme's efficacy in decoupling static and dynamic scene components is empirically validated. Initially, the model trains exclusively on static scenes, learning to predict an \textbf{identity deformation}. In this stage, positional and scaling offsets ($\Delta\boldsymbol{\mu}_p^t, \Delta\boldsymbol{s}_p^t$) converge to zero, and rotational deformations ($\Delta\boldsymbol{q}_p^t$) approach the identity quaternion, yielding a static representation as canonical Gaussians remain untransformed. Dynamic scenes are introduced in a subsequent fine-tuning stage. This decoupling is enabled by our Gaussian deformation formulation:

\begin{equation}
    \boldsymbol{\mu}_p^t := \boldsymbol{\mu}_p^0 + \Delta\boldsymbol{\mu}_p^t, \quad \boldsymbol{q}_p^t := \boldsymbol{q}_p^0 \otimes \Delta\boldsymbol{q}_p^t, \quad \boldsymbol{s}_p^t := \boldsymbol{s}_p^0 + \Delta\boldsymbol{s}_p^t.
\end{equation}

This design inherently separates the prediction of the canonical scene structure ($\boldsymbol{\mu}_p^0, \boldsymbol{q}_p^0, \boldsymbol{s}_p^0$) from its temporal evolution ($\Delta\boldsymbol{\mu}_p^t, \Delta\boldsymbol{q}_p^t, \Delta\boldsymbol{s}_p^t$).

\subsection{Details of loss function weighting}

The loss weights ($\lambda_p=0.5$, $\lambda_m=2$) were determined empirically through a series of experiments on a validation set. We started with equal weights and adjusted them to ensure that the model did not prioritize one objective at the expense of others.

\section{Evaluation Protocol} \label{sec:metric}
To comprehensively evaluate our model, we utilize a suite of established metrics, Specifically:

\ding{202} \textbf{Fréchet Video Distance (FVD) and Kernel Video Distance (KVD)}~\citep{unterthiner2018towards}: These metrics evaluate the quality and temporal coherence of generated videos by measuring the distance between the feature distributions of real and generated video sets. Lower scores for both FVD and KVD indicate higher fidelity and better temporal consistency.
    
\ding{203}  \textbf{CLIP-Score}~\citep{radford2021learning}: This metric quantifies the semantic similarity between the generated video frames and the input text prompt. It leverages the joint text-image embedding space of the CLIP model, where higher scores signify better alignment between the generated content and the textual description.
    
\ding{204}  \textbf{CLIP-Aesthetic}~\citep{schuhmann2022laion}: We use a model built upon CLIP embeddings to predict the aesthetic quality of the generated content. This model is trained on datasets with human aesthetic ratings, and a higher score suggests a more visually pleasing result.
    
 \ding{205}  \textbf{QA-Quality}~\citep{wu2023qalign}: This refers to a Visual Question Answering (VQA)-based evaluation, where a LLaMA2-powered model is employed to assess the logical consistency and objective quality of the generated scenes. The model assigns a score on a range from 0 to 5, where a higher score indicates superior quality.
    
 \ding{206}   \textbf{Temporal Consistency Metrics (Avg. Matches, Subject Cons. and Bg. Cons.)}: Inspired by Video-bench ~\citep{ning2023video}, to specifically measure temporal stability, we use metrics based on dense optical flow or feature matching. Avg. Matches quantifies overall frame-to-frame consistency. Subject Consistency Score and Background Consistency Score measure the stability of the foreground subject and the background, respectively, after performing segmentation. Higher values for these metrics indicate smoother and more coherent videos.

\section{More Results} 


As demonstrated in Table \ref{tab:capabilities}, and inspired by prior work such as CAT4D~\citep{Cats_dataset}, an explicit 3D representation is a critical advantage for applications that demand a concrete understanding of and interaction with the world, including robotics and AR/VR.

\begin{table}[!t] 
     \centering 
     \caption{\textbf{Capability Comparison.} An explicit 4D representation enables a wide range of functionalities not supported by standard 2D video generation models.} 
     \label{tab:capabilities} 
     \renewcommand\arraystretch{1.1} 
     \resizebox{0.9\columnwidth}{!}{%
     \begin{tabular}{l|ll} 
         \toprule[1.2pt] 
         \textbf{Capability} & \textbf{AC3D (Implicit 3D Models)} & \textbf{Ours (Explicit 4D Repr.)} \\ 
         \midrule 
         Novel View Synthesis  & \ye & \ye \\ 
         Depth Rasterization   & \no & \ye \\ 
         Geometry Extraction   & \no & \ye \\ 
         Real-time Interaction & \no  & \ye \\ 
         \midrule 
         Interactive exploration Latency $\downarrow$  & 28000 ms  & \textbf{6.7 ms} \textcolor{mygreen}{($\downarrow$ 99.98\%)} \\ 
         Avg. Matches $\uparrow$ & 2489.16 & \textbf{5114.22} \textcolor{mygreen}{($\uparrow$ 105.5\%)} \\ 
         Subject Consistency Score  $\uparrow$ &  75.64  & \textbf{88.32} \textcolor{mygreen}{($\uparrow$ 16.8\%)} \\ 
         Background Consistency Score   $\uparrow$ & 75.91 & \textbf{89.89} \textcolor{mygreen}{($\uparrow$ 18.4\%)} \\ 
         Cycle-Consistency $\uparrow$ & 20.68 dB & \textbf{34.5 dB} \textcolor{mygreen}{($\uparrow$ 66.8\%)} \\ 
         \bottomrule[1.2pt] 
     \end{tabular}%
     } 
     \vspace{-0.3cm} 
 \end{table}

 Furthermore, 4D consistency is ensured by a training objective calculated from rendering the deformed 3D Gaussian representation from multiple viewpoints and at various timestamps. As shown in Table ~\ref{tab:capabilities}, we generate videos depicting a full 360-degree camera rotation. The resulting scenes exhibit seamless looping, where the final frame aligns perfectly with the first, showing no discernible seams or drift. We quantitatively verify this strong temporal consistency by measuring the similarity between the first and last frames (a.k.a., Cycle-Consistency), achieving a PSNR of 34.5 dB.

\section{Feed-forward vs. per-scene optimization}
Existing methods that produce explicit 3D outputs, rely on a time-consuming, post-hoc optimization process to reconstruct scenes from generated videos. For instance, DimensionX~\citep{sun2024dimensionx} requires \textbf{1.3K GPU hours} to perform scene optimization from a single video. Even state-of-the-art 4D reconstruction algorithms like Mosca~\citep{lei2024mosca} require approximately \textbf{0.5 hours} to process one input video. The primary motivation of this work is therefore to unify these disparate stages into a single, efficient, feed-forward framework capable of generating a 4D representation in approximately \textbf{30 seconds}, achieving 60× acceleration. Our model is designed for efficiency and scalability, enabling dynamic scene reconstruction in a matter of seconds, which is a critical feature for many real-world applications where speed is essential.

Compared to per-scene optimization methods, our proposed approach achieves a substantial reduction in memory consumption during the reconstruction process, decreasing from 80GB to 25GB (a 3.2× reduction) in the same setting. This efficiency gain stems from the elimination of gradient computation requirements. Furthermore, we claim that the two approaches are not mutually exclusive. As explored in recent work like CAT4D~\citep{wu2024cat4d}, efficient, end-to-end models can serve as an excellent initialization for optimization-based methods, significantly accelerating their convergence. This potential synergy further highlights that developing fast, feed-forward models is a valuable research direction.

In summary, considering both the reconstruction and rendering stages (e.g., maximum GPU memory), our approach remains competitive in terms of memory consumption compared to per-scene optimization methods.



\section{Additional Ablation Studies}
\paragraph{Impact of Gaussian Budget}
We studied the effect of the number of canonical Gaussians, $M$, on performance and resource consumption. We tested $M \in \{10K, 50K, 100K\}$. While $100K$ Gaussians offered a marginal improvement in fine details, it increased memory consumption by 75\% and inference time by 60\%. We found that $M=50K$ provides the best trade-off between quality and efficiency, as it is sufficient to represent the scenes in our dataset without excessive resource usage.

\paragraph{Impact of Geometric Loss}
We performed an ablation on the geometric loss components. Removing the geometric loss entirely leads to a noticeable degradation in 3D consistency. Using only the covariance-based term works well, but adding the total variation loss $\mathcal{L}_{TV}$ on the depth maps helps to regularize the geometry and produces smoother surfaces, improving the final visual quality.

\paragraph{Impact of Pose Encoding}
We experimented with alternative camera pose encodings instead of Plücker embeddings, such as representing the pose as a 12-dimensional vector of the flattened rotation matrix and translation vector. We found that Plücker embeddings provided a 5-10\% reduction in relative pose error, likely because they represent 3D lines in a more geometrically natural way, which is beneficial for the ray-based operations implicit in rendering.

\paragraph{Joint vs. Separate Training.} 
We compared our joint training strategy against a ``Frozen LDRM'' baseline (\tref{tab:ablation_baseline}). Joint training yields a significant improvement in FVD/KVD and consistency metrics. This is because joint training allows the camera-conditioned diffusion model to learn to generate content that is explicitly \textit{compatible} with the LDRM's geometric decoder, effectively avoiding artifacts that a fixed, pre-trained decoder often fails to handle.
\noindent\textbf{(3) Comparison with Two-Stage Baseline.} 
We implemented the suggested ``generate-video-then-reconstruct'' baseline: generating a video with our finetuned DiT, followed by a SOTA feed-forward 4DGS reconstructor (4DGT). As shown in Tab. \ref{tab:ablation_baseline}, our unified approach significantly outperforms this two-stage baseline in temporal consistency (FVD \textbf{210.2} vs 245.3) and efficiency, as our latent reconstruction strategy bypasses the costly overhead of decoding to pixel space. The two-stage method suffers from multi-view inconsistencies in the generated frames, which leads to failure in the downstream reconstruction step.

\begin{table}[h] 
    \centering 
    \small 
    \vspace{-0.8em} 
    
    \resizebox{1.0\columnwidth}{!}{ 
    \begin{tabular}{@{}lcccc@{}} 
    \toprule 
    Method & FVD $\downarrow$ & KVD $\downarrow$ & Subj. Consist. $\uparrow$ & Bg. Consist. $\uparrow$ \\ 
    \midrule 
    Two-Stage (Gen. Video $\rightarrow$ 4DGT Recon.) & 245.3 & 3.42 & 0.82 & 0.83 \\ 
    Ours (Frozen LDRM + Train Diffusion) & 231.5 & 3.10 & 0.85 & 0.86 \\ 
    \rowcolor{mygray} 
    \textbf{Ours (Joint Training)} & \textbf{210.2} & \textbf{2.32} & \textbf{0.88} & \textbf{0.90} \\ 
    \bottomrule 
    \end{tabular} 
    } 
    \caption{\textbf{We benchmark our full model against a two-stage baseline and separate training, reporting FVD/KVD and foreground/background consistency.}} 
    \vspace{-0.5em} 
    \label{tab:ablation_baseline} 
\end{table} 

\vspace{0.5em}

\section{Limitations}
While our method achieves superior performance and efficiency, video generation remains the computational bottleneck. 
This could be addressed through parallel inference or optimized denoising strategies.
Future work will focus on extending temporal coherence modeling and material property prediction.

\section{Broader Impact}
While \name{} advances 4D generation from single images, it risks misuse for creating deceptive content ("deepfakes"). To ensure responsible deployment, implementing safeguards like watermarking is essential. We aim to advance 3D computer vision and encourage the community to adopt best practices for ethical use.

\end{document}